\documentclass[
twocolumn,
]{ceurart}

\usepackage{listings}
\usepackage{graphicx} 

\usepackage{framed}
\usepackage{xcolor}
\usepackage{multirow}
\usepackage{multicol}
\usepackage{enumitem}
\usepackage{array}
\usepackage{booktabs}  
\usepackage{tabularx}

\newcounter{customboxcounter}
\newenvironment{custombox}[1]{
  \refstepcounter{customboxcounter}
  \MakeFramed{\FrameRestore}
  \noindent\textbf{#1} 
  \par\vskip6pt}{
  \endMakeFramed}

\begin{document}

\copyrightyear{2025}
\copyrightclause{Copyright for this paper by its authors.
  Use permitted under Creative Commons License Attribution 4.0
  International (CC BY 4.0).}

\conference{7th Workshop on Automated Semantic Analysis of Information in Legal Text, 
16 June 2025, Chicago, IL}

\title{Measuring Faithfulness and Abstention: An Automated Pipeline for Evaluating LLM-Generated 3-ply Case-Based Legal Arguments} 


\author[1]{Li Zhang}[%
orcid=0000-0003-0375-1793,
email=liz239@pitt.edu,
]
\cormark[1]

\author[1]{Morgan Gray}[%
orcid=0000-0002-3800-2103,
email=mag454@pitt.edu,
]

\author[2]{Jaromir Savelka}[%
orcid=0000-0002-3674-5456,
email=jsavelka@andrew.cmu.edu,
]

\author[1]{Kevin D. Ashley}[%
orcid=0000-0002-5535-0759,
email=ashley@pitt.edu,
]

\address[1]{Intelligent Systems Program, University of Pittsburgh,
  Pittsburgh, Pennsylvania, USA}

\address[2]{School of Computer Science, Carnegie Mellon University,
  Pittsburgh, Pennsylvania, USA}

\cortext[1]{Corresponding author.}

\begin{abstract}
Large Language Models (LLMs) demonstrate potential in complex legal tasks like argument generation, yet their reliability remains a concern. Building upon pilot work assessing LLM generation of 3-ply legal arguments using human evaluation, this paper introduces an automated pipeline to evaluate LLM performance on this task, specifically focusing on faithfulness (absence of hallucination), factor utilization, and appropriate abstention. We define hallucination as the generation of factors not present in the input case materials and abstention as the model's ability to refrain from generating arguments when instructed and no factual basis exists. Our automated method employs an external LLM to extract factors from generated arguments and compares them against the ground-truth factors provided in the input case triples (current case and two precedent cases). We evaluated eight distinct LLMs on three tests of increasing difficulty: 1) generating a standard 3-ply argument, 2) generating an argument with swapped precedent roles, and 3) recognizing the impossibility of argument generation due to lack of shared factors and abstaining. Our findings indicate that while current LLMs achieve high accuracy (over 90\%) in avoiding hallucination on viable argument generation tests (Tests 1 \& 2), they often fail to utilize the full set of relevant factors present in the cases. Critically, on the abstention test (Test 3), most models failed to follow instructions to stop, instead generating spurious arguments despite the lack of common factors. This automated pipeline provides a scalable method for assessing these crucial LLM behaviors, highlighting the need for improvements in factor utilization and robust abstention capabilities before reliable deployment in legal settings. Project page: \href{https://lizhang-aiandlaw.github.io/An-Automated-Pipeline-for-Evaluating-LLM-Generated-3-ply-Case-Based-Legal-Arguments/}{Link}.
\end{abstract}


\maketitle

\section{Introduction}

Large language models (LLMs) have demonstrated remarkable capabilities across various domains, including legal analysis and argumentation \cite{ariai2024natural,katz2023natural,gray2025generating}. Their potential to streamline legal research, draft documents, and even generate arguments offers significant efficiency gains. However, their tendency to hallucinate facts or generate plausible but unsupported statements poses significant risks in legal applications, where accuracy and reliability are of utomost importance \cite{bommasani2021opportunities,ji2023survey}. Misguided decisions, ethical concerns, and even professional sanctions can result from relying on inaccurate AI-generated legal content \cite{de2024artificial,avery2023chatgpt}.

A critical challenge lies in ensuring the factual accuracy and appropriate reasoning behavior of LLMs when tasked with generating case-based legal arguments. Pilot work involving human evaluation of LLM-generated 3-ply arguments (plaintiff's argument citing precedent 1, defendant's counterargument distinguishing precedent 1 and citing precedent 2, plaintiff's rebuttal distinguishing precedent 2) indicated that while LLMs can produce structurally coherent arguments, their factual grounding and adherence to constraints can be problematic \cite{gray2025generating}. Specifically, LLMs may hallucinate, i.e., introduce factual elements (represented as `factors' in case-based reasoning) not present in the source materials. Furthermore, they may fail to follow instructions appropriately, particularly negative constraints such as abstaining from generating an argument when the provided cases lack a sufficient factual basis for comparison. Existing evaluation methods often focus on general capabilities \cite{guha2024legalbench,li2024lexeval,fei2023lawbench} but lack fine-grained metrics to assess these specific failure modes in the context of factor-based legal argumentation.

To address this gap, we introduce an automated pipeline for evaluating LLM performance in generating 3-ply, factor-based legal arguments. This pipeline specifically targets the assessment of hallucination, factor utilization (the extent to which relevant, available factors are used), and appropriate abstention. The core of our approach involves using an external LLM to analyze the arguments generated by the models under test, extracting the factors cited within them. These extracted factors are then compared against the ground-truth factors present in the input case materials to compute quantitative metrics for faithfulness and completeness.

The development of this automated evaluation pipeline enables a targeted assessment of LLM behavior in generating factor-based arguments. To guide this assessment, we pose the following research questions (RQs):

\begin{itemize}
    \item \textbf{RQ1:} To what extent do LLMs exhibit measurable errors, specifically hallucination (citing non-existent factors) and incomplete factor utilization (omitting relevant available factors), when tasked with generating 3-ply case-based arguments from factor-represented inputs?
    
    \item \textbf{RQ2:} How effectively do LLMs adhere to instructions to abstain from argument generation when presented with input cases lacking common factors, and what is their propensity to generate spurious arguments under such conditions?
    
    \item \textbf{RQ3:} Can the proposed automated evaluation metrics effectively quantify distinct error types (hallucination, incomplete utilization, spurious generation) and successfully reveal performance variations across different LLMs and varying levels of task complexity?
\end{itemize}

\noindent The main contributions of this paper are: an automated evaluation pipeline specifically designed for assessing LLM-generated, factor-based legal arguments; novel metrics targeting hallucination, factor utilization, and abstention behavior in this context; an empirical evaluation of eight distinct LLMs (including open-source and proprietary models of varying sizes) on three argumentation tasks with increasing difficulty; and insights into the specific weaknesses of current LLMs regarding factual grounding and instruction following in legal argument generation.

The remainder of this paper is structured as follows: Section 2 discusses related work. Section 3 details the methodology of our automated evaluation pipeline. Section 4 describes the experimental setup, including the dataset, tasks, and models. Section 5 presents the results of our evaluation. Section 6 provides a qualitative error analysis. Section 7 concludes the paper. Finally, Section 8 acknowledges the limitations and suggests future work.

\section{Background and Related Work}

\subsection{LLMs in the Legal Domain}
Recent advances in LLMs, from open-source models such as Llama models \cite{touvron2023llama} to proprietary systems such as GPT models \cite{achiam2023gpt}, have demonstrated remarkable capabilities in natural language understanding and generation. This has spurred significant interest in their application within the legal domain, ranging from legal research assistance and contract analysis to case outcome prediction and automated legal document drafting \cite{ariai2024natural}. While promising, the reliable deployment of these models requires careful consideration of their limitations, particularly concerning factual accuracy.

\subsection{Computational Argumentation and Case-Based Reasoning}
Computational models of legal argument, particularly those employing case-based reasoning (CBR), provide a foundation for analyzing and generating legal arguments. Early work pioneered the use of 'factors'---stereotypical fact patterns relevant to legal claims---in US trade secrets law, developing systems like HYPO that analyze and compare cases based on shared and distinguishing factors \cite{ashley1990}. Subsequent systems like CATO introduced factor hierarchies \cite{aleven1997teaching}, while others integrated rule-based and case-based approaches \cite{rissland1991cabaret} or focused on predicting outcomes \cite{bruninghaus2003predicting} and incorporating legal values \cite{grabmair2017predicting}. Formal models of precedential constraint based on factors have also been developed \cite{horty2021modifying}. Factors provide a structured representation suitable for evaluating the factual basis of arguments, as employed in this study.

\subsection{Argument Generation with LLMs}
Beyond general legal tasks, researchers are exploring the specific capability of LLMs to generate arguments. Some work has shown LLMs can assist humans in identifying legal factors \cite{gray2024empirical,savelka2023unreasonable} or generate factor-based arguments in a structured manner \cite{gray2025generating}. However, the practical utility of such generated arguments hinges on their factual accuracy and logical coherence. This paper focuses on evaluating these aspects rather than proposing new generation methods.

\subsection{Hallucination in LLMs}
A significant challenge for LLMs is hallucination—the generation of content that is factually incorrect or unsupported by the provided input or established knowledge \cite{ji2023survey,yadkori2024believe}. Hallucinations can manifest as contradictions with the input prompt, conflicts with provided context, or deviations from real-world facts \cite{zhang2023siren}. In high-stakes domains like law, where precision and truthfulness are critical \cite{bommasani2021opportunities}, hallucination represents a major barrier to adoption. Mitigation strategies often involve techniques like chain-of-thought prompting \cite{dhuliawala2023chain} or retrieval-augmented generation (RAG) to ground responses in external sources \cite{vu2023freshllms}. Our work focuses on reliably detecting hallucination in the specific context of factor-based legal arguments.

\subsection{Evaluation Metrics for Generated Text}
Standard metrics for evaluating generated text, such as ROUGE \cite{lin2004rouge}, BLEU \cite{papineni2002bleu}, and BERTScore \cite{zhang2019bertscore}, primarily measure surface-level similarity or semantic overlap with reference texts. While useful for tasks like summarization or translation, they are often insufficient for assessing factual accuracy, logical consistency, or adherence to constraints in complex generation tasks like legal argumentation \cite{fei2023lawbench}. Some legal benchmarks exist \cite{guha2024legalbench,li2024lexeval}, but metrics specifically tailored to evaluate faithfulness and abstention in factor-based reasoning remain underdeveloped. Our work aims to fill this gap by proposing automated metrics focused on these critical aspects.

\subsection{Instruction Following in LLMs}
The ability of LLMs to accurately follow complex instructions is crucial for their reliable use. Research has shown that while LLMs are increasingly capable of adhering to instructions, they can struggle with nuanced or complex constraints, particularly negative constraints (e.g., ``do not generate X if condition Y is met'') or situations requiring implicit recognition of task impossibility \cite{wen2024know,feng2024don}. Failure to follow instructions, such as the requirement to abstain from generating an argument when no factual basis exists, is one of the key failure modes investigated in this study.

\section{Methodology}

Our work employs a pipeline designed to automatically generate, assess, and score LLM performance on a structured legal argument generation task. This pipeline consists of several stages: scenario generation, argument generation by the models under test, automated factor extraction from the generated arguments, and quantitative scoring based on comparison with ground-truth inputs.

\subsection{Task Definition: 3-Ply Argument Generation}
\begin{figure}[h]
    \includegraphics[width=\linewidth]{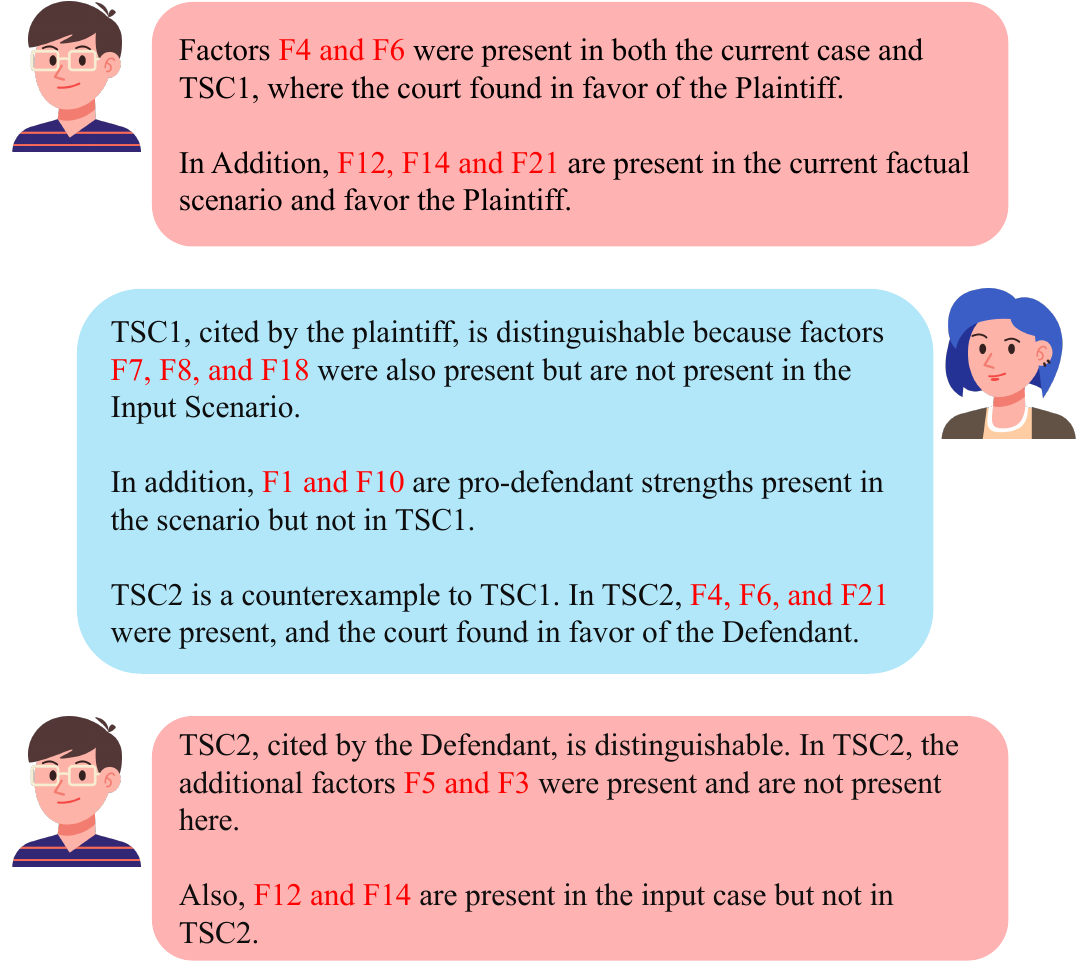}
    \caption{Three-ply Legal Argument Generation Scheme}
    \label{fig:3ply-scheme}
\end{figure}

The core task requires the LLM to generate a 3-ply legal argument within the U.S. trade secret law domain, following the structure established by \citet{ashley1990}. Given a current fact situation (the `current case') represented by factors, and two precedent trade secret misappropriation cases (TSC1 and TSC2), also represented by factors, the LLM must perform three steps: First, act as Plaintiff and argue for victory by citing TSC1 or TSC2 as an analogous precedent, focusing on shared pro-plaintiff factors. Second, act as Defendant, responding by distinguishing Plaintiff's cited case (highlighting differential factors) and citing the other precedent case as a counter-example favouring the defendant, focusing on shared pro-defendant factors. Third, act again as Plaintiff for a rebuttal, distinguishing the Defendant's cited case and emphasizing factors that differentiate the current case from the Defendant's precedent. This argumentative structure is illustrated in Figure \ref{fig:3ply-scheme}.

\subsection{Factor-Based Case Representation}
Cases are represented using a standardized set of 26 legal factors pertinent to U.S. trade secret misappropriation law, derived from foundational work in legal AI \cite{ashley1990, ashley2017artificial}. These factors encapsulate key factual aspects, such as circumstances surrounding disclosure, security measures implemented, characteristics of the information, and relevant employee conduct. Each factor is designated as typically favouring either the plaintiff (P) or the defendant (D). For instance, a case might be represented textually as: 

\begin{quote}
\texttt{[Case Name] [Outcome] [Factors: F1 Disclosure-in-negotiations (D), F4 Agreed-not-to-disclose (P), F6 Security-measures (P)]}
\end{quote}

\noindent This structured factor representation facilitates objective comparison between cases based on shared and distinguishing factors, providing the essential ground truth for our subsequent automated evaluation metrics.

\subsection{Argument Generation and Evaluation Pipeline}
\begin{figure}[htbp]
    \centering
    \includegraphics[width=\linewidth]{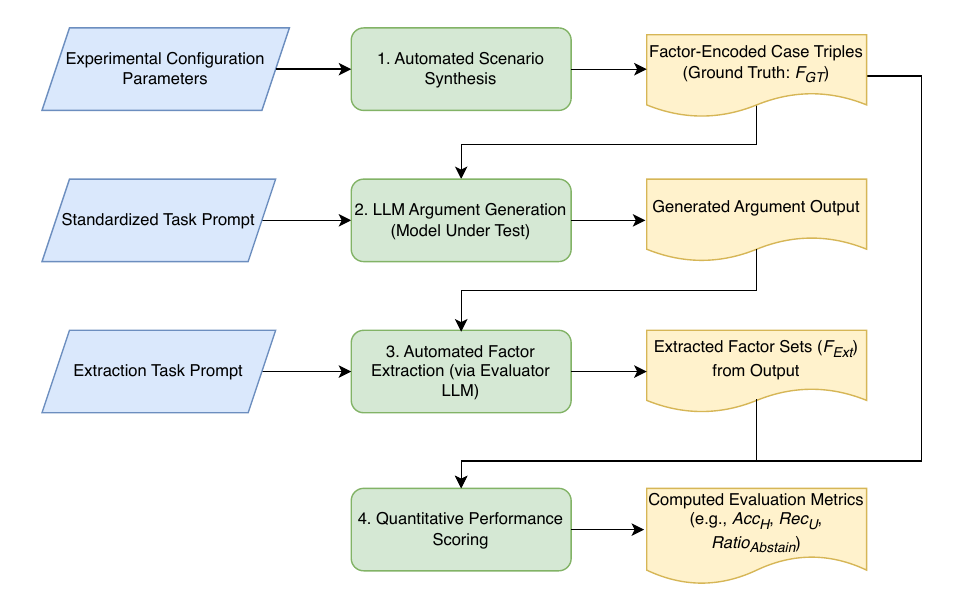}
    \caption{Overall Automated Evaluation Pipeline Flowchart}
    \label{fig:pipeline_flowchart}
\end{figure}

The evaluation process follows a defined pipeline. First, legal scenarios (case triples) are generated according to specific criteria (detailed in Section 4.1). Second, each model under test receives these scenarios within a structured prompt (Section 4.2) and generates the 3-ply argument. Third, an automated process extracts the factors cited within the generated argument text (Section 3.4.1). Finally, these extracted factors are compared against the ground-truth factors from the input scenario to calculate performance metrics (Sections 3.4.2-3.4.4). The overall process is depicted in Figure~\ref{fig:pipeline_flowchart}.

The core argument generation step takes the factor-represented case triple as input and invokes the chosen model with specific parameters (Section 4.4) to produce the 3-ply textual argument. The LLM output, along with metadata, is logged for subsequent analysis.

\subsection{Automated Metrics Definition}
To evaluate the generated 3-ply arguments quantitatively, we developed automated metrics focused on faithfulness (absence of hallucination), factor utilization (completeness), and adherence to abstention instructions. These metrics rely on comparing the factors cited within the generated argument against the ground-truth factors present in the input cases.

\textbf{3.4.1 Factor Extraction:} An external, high-capability LLM (\texttt{GPT-4.1}) serves as an automated evaluator. For each generated 3-ply argument produced by a model under test, this evaluator LLM is prompted to analyze the argument text. Its task is to identify and extract the specific sets of factors that the model under test asserted existing in each case in the triples (Current Case - CC, TSC1, TSC2). Let $F_{Ext, c}$ be the set of factors extracted by the evaluator for case $c \in \{CC, TSC1, TSC2\}$. Similarly, let $F_{GT, c}$ be the set of actual ground-truth factors present in the input for case $c$.

\textbf{3.4.2 Hallucination Metric:} Hallucination is operationally defined as the assertion by the model under test that a specific factor exists in a specific case when, according to the ground-truth input for that case, it is not present. We quantify this by summing the hallucinations across all three cases and normalizing by the total number of ground-truth factors in the input triple. The \textit{Hallucination Accuracy} ($Acc_H$) is calculated as:
\[ Acc_H = \left(1 - \frac{N_H}{N_{GT}}\right) \times 100\% \]
Here, $N_H$ is the total count of hallucinated factors across the three cases:
\[ N_H = \sum_{c \in \{CC, TSC1, TSC2\}} |\{f \in F_{Ext, c} \mid f \notin F_{GT, c}\}| \]
And $N_{GT}$ represents the total count of factors across the three ground-truth input cases (sum of factors per case, not unique factors):
\[ N_{GT} = \sum_{c \in \{CC, TSC1, TSC2\}} |F_{GT, c}| \]
A higher $Acc_H$ indicates greater faithfulness, meaning the argument relies less on unsupported factual assertions specific to each case.

\textbf{3.4.3 Factor Utilization Metric:} Factor utilization assesses how comprehensively the model under test mentions the available ground-truth factors for the specific cases they belong to. We compute \textit{Factor Utilization Recall} ($Rec_U$) by summing the correctly identified factors for each case and normalizing by the total number of ground-truth factors.
\[ Rec_U = \left(\frac{N_U}{N_{GT}}\right) \times 100\% \]
where $N_U$ is the total count of utilized ground-truth factors correctly mentioned for their respective cases across the triple:
\[ N_U = \sum_{c \in \{CC, TSC1, TSC2\}} |F_{Ext, c} \cap F_{GT, c}| \]
$N_{GT}$ is defined as above. A higher $Rec_U$ indicates that the generated argument incorporates more of the factual elements provided in the input, correctly associating them with their respective cases.

\textbf{3.4.4 Abstention Metric:} Test 3 specifically tests the model's ability to abstain when argument generation is impossible due to a lack of shared factors. The primary measure for this test is the \textit{Abstention Ratio} ($Ratio_{Abstain}$). Let $N_{SA}$ be the number of successfully executed abstentions and $N_{TA}$ be the total number of test triples requiring abstention. The ratio is calculated as:
\[ Ratio_{Abstain} = \left(\frac{N_{SA}}{N_{TA}}\right) \times 100\% \]
A higher $Ratio_{Abstain}$ indicates better adherence to instructions to abstain. For this test, we also report Hallucination Accuracy ($Acc_H$) to characterize the nature of the arguments generated when models failed to abstain (Section 5.3).

\subsection{Rationale for External LLM-based Evaluation}
Employing a highly capable LLM (\texttt{GPT-4.1}) for the factor extraction step (Section 3.4.1) provides substantial advantages in scalability and consistency compared to manual annotation across potentially hundreds or thousands of generated arguments. Although the evaluator LLM is not infallible, our spot checks indicated high accuracy in identifying factor mentions within the generated text structures. This automated approach enables large-scale, reproducible evaluation across numerous models and experimental conditions. Potential limitations associated with this method are acknowledged in Section 8.

\section{Experimental Design}

\subsection{Dataset Generation and Structure}
The dataset used in this study was synthetically generated using a custom tool designed to create controlled case triples for evaluating specific argumentation phenomena within the U.S. trade secret domain. Each generated triple includes a factor-represented current case, a potential plaintiff precedent (TSC1), and a potential defendant counter-precedent (TSC2).

The generation process allows for specifying several parameters, including the number of cases, the complexity level which controlls the number of factors per case, typically ranging from complexity-1 to complexity+1, and, crucially, the scenario `mode'. We generated data across three distinct modes, each designed to test different facets of LLM reasoning and instruction following:
\begin{itemize}[noitemsep, topsep=0pt]
    \item \textbf{Arguable Sets:} These triples contain sufficient overlapping factors between the current case and the respective precedents (TSC1 for plaintiff, TSC2 for defendant), with aligned outcomes, facilitating standard 3-ply argument generation. These are used for Test 1.
    \item \textbf{Reordered Sets:} In these triples, common factors exist, but the typical roles based on outcomes are reversed (TSC1 favors Defendant, TSC2 favors Plaintiff). These are used for Test 2, primarily testing robustness to the reordered precedent cases compared to Test 1.
    \item \textbf{Non-arguable Sets:} These triples are constructed such that there are no common factors between the current case and either TSC1 or TSC2. These are specifically designed for Test 3 to evaluate the model's ability to recognize the impossibility of argument generation and abstain as instructed.
\end{itemize}
For this study, we generated sets of 30 triples with complexity of 12, for each of the `Arguable', `Reordered', and `Non-arguable' modes, resulting in a total dataset of 90 triples used across the three experimental tests. Example case structures for each mode are illustrated in Table \ref{tab:case_set_examples}. This structured dataset allows for targeted testing of baseline argument generation (Arguable), adherence to specific instructions under potentially confusing conditions (Reordered/Swapped Roles), and the crucial ability to recognize factual impossibility and follow abstention instructions (Non-arguable).

\begin{table*}[htbp]
\caption{Examples of Different Dataset Scenario Modes}
\label{tab:case_set_examples}
\centering
\small
\begin{tabular}{p{0.15\textwidth}p{0.25\textwidth}p{0.25\textwidth}p{0.25\textwidth}}
\hline
\textbf{Mode} & \textbf{Current Case} & \textbf{TSC1} & \textbf{TSC2} \\
\hline
Arguable & F4: Agreed-not-to-disclose (P)*\newline F5 Agreement-not-specific (D)†\newline F23 Waiver-of-confidentiality (D)  & outcome: Plaintiff\newline F2 Bribe-employee (P)\newline F4: Agreed-not-to-disclose (P)*\newline F16 Info-reverse-engineerable (D) & outcome: Defendant\newline F2 Bribe-employee (P)\newline F5: Agreement-not-specific (D)†\newline F12: Outsider-disclosures-restricted (P)\\
\hline
Reordered & F4: Agreed-not-to-disclose (P)†\newline F5 Agreement-not-specific (D)*\newline F23 Waiver-of-confidentiality (D)  & outcome: Defendant\newline F2 Bribe-employee (P)\newline F5: Agreement-not-specific (D)*\newline F12: Outsider-disclosures-restricted (P) & outcome: Plaintiff\newline F2 Bribe-employee (P)\newline F4: Agreed-not-to-disclose (P)†\newline F16 Info-reverse-engineerable (D)  \\
\hline
Non-arguable & F6: Security-measures (P)\newline F22: Invasive-techniques (P) & outcome: Plaintiff\newline F1: Disclosure-in-negotiations (D)\newline F27: Disclosure-in-public-forum (D) & outcome: Defendant\newline F16: Info-reverse-engineerable (D)\newline F24: Info-obtainable-elsewhere (D) \\
\hline
\end{tabular}
\vspace{-2mm} 
\begin{minipage}{\textwidth}
\scriptsize Notes: Common factors between Current Case and TSC1 marked with *; common factors between Current Case and TSC2 marked with †.
\end{minipage}
\end{table*}

\subsection{Core Prompt Structure}
For each test, the LLMs under evaluation were provided with a structured prompt designed to give sufficient context and clear instructions. The prompt included a description of the 3-ply argument test (as the scheme shown in Figure~\ref{fig:3ply-scheme}), relevant background on trade secret misappropriation law, and the factor representations for the specific input cases (current case, TSC1, TSC2). Crucially, the prompt also contained explicit instructions regarding the desired output format and an abstention condition: it stated that if no common factors could be found to support an analogy for a given ply, the model should output a specific phrase (e.g., ``Cannot generate argument due to lack of common factors'') and stop processing that ply, rather than fabricating an argument. This abstention instruction was particularly relevant for evaluating performance on Test 3. An example of the full prompt structure is provided in Appendix \ref{app:example_prompt} and \ref{app:factor_extraction_prompt}.

\subsection{Curriculum for Testing}
We evaluated the selected LLMs on three distinct tests, leveraging the different modes of our generated dataset:
\begin{itemize}[noitemsep, topsep=0pt]
    \item \textbf{Test 1 on Arguable Sets: Standard Argument Generation.} Using the `Arguable' case triples, models generated the standard 3-ply argument (Plaintiff cites TSC1, Defendant cites TSC2, Plaintiff rebuts TSC2). This test assesses baseline performance regarding hallucination and factor utilization when arguments are factually supported.
    \item \textbf{Test 2 on Reordered Sets: Swapped Precedent Roles.} Also using the `Arguable' triples, this test required models to perform the 3-ply argument but with the order of TSC1 and TSC2 swapped (TSC1 favoring the Defendant, TSC2 favoring the Plaintiff). Critically, models were not given the precedent name to cite; instead, they had to select the appropriate precedent (TSC1 or TSC2) by analyzing which case's outcome supported their argumentative goal.
    \item \textbf{Test 3 on Non-arguable Sets: Abstention Test.} Utilizing the `Non-arguable' case triples, models were prompted to generate the standard 3-ply argument. The expected correct behavior, however, was for the model to recognize the lack of common factors required for analogical reasoning and follow the explicit instruction to abstain from generating spurious arguments for triple. This test directly probes the ability to identify task impossibility and adhere to negative constraints.
\end{itemize}
These tests present progressively challenging scenarios designed to probe different aspects of LLM reliability in legal argument generation.

\subsection{Models Evaluated}
We selected eight distinct LLMs, representing a range of sizes, architectures, and access types (open-source and commercial). The models evaluated were:
\begin{itemize}[noitemsep, topsep=0pt]
    \item \textit{GPT-4o} (OpenAI)
    \item \textit{GPT-4o-mini} (OpenAI)
    \item \textit{Llama-3-70B-8192} (Meta)
    \item \textit{Llama-3-8B-8192} (Meta)
    \item \textit{Llama-4-Maverick-17B-128e-instruct} (Meta)
    \item \textit{Llama-4-Scout-17B-16e-instruct} (Meta)
    \item \textit{DeepSeek-R1-Distill-Llama-70B} (DeepSeek)
    \item \textit{Qwen-QWQ-32B} (Alibaba)
\end{itemize}
This selection aimed to cover a spectrum of capabilities, including models ranging from comparatively small (Llama-3-8B) to large (GPT-4o, Llama-3-70B), proprietary and open-source options, Mixture-of-Experts architectures (Llama-4 variants), and models optimized for reasoning tasks (Qwen-QWQ-32B, DeepSeek-R1-Distill-Llama-70B). These models were accessed via their respective APIs at the time of experimentation (May 2025).

\subsection{Implementation Details}
To ensure comparability across models and tests, consistent generation parameters were employed for all LLM invocations within the argument generation pipeline. We used a \texttt{temperature} setting of 0 since the task is focused on expected correct output. A \texttt{max\_tokens} limit of 500 was set, which proved sufficient for the typical length of a 3-ply argument structure (for reasoning models, the limit was set as 5,000). Other standard parameters included \texttt{top\_p}=1, \texttt{frequency\_penalty}=0, and \texttt{presence\_penalty}=0. The factor extraction step performed by the external evaluator LLM (\texttt{GPT-4.1}) also utilized fixed deterministic settings to ensure consistency in the evaluation process itself.

\section{Results}

We analyze the performance of the eight evaluated LLMs across the three distinct tests using the automated metrics for Hallucination Accuracy ($Acc_H$, Section 3.4.2), Factor Utilization Recall ($Rec_U$, Section 3.4.3), and Abstention Ratio ($Ratio_{Abstain}$, Section 3.4.4), as defined previously. The results are presented separately for each test type: Test 1 (Arguable), Test 2 (Reordered), and Test 3 (Non-arguable).

\begin{table*}[!htbp]
\caption{LLM Hallucination Accuracy ($Acc_H$, \% Across Tests)}
\label{tab:accuracy_results}
\centering
\footnotesize
\setlength{\tabcolsep}{6pt} 
\begin{tabular}{lccc}
\toprule
\textbf{Model} & \textbf{Test 1 (Arguable)} & \textbf{Test 2 (Reordered)} & \textbf{Test 3 (Non-arguable)} \\
\midrule
Llama-3-8B-8192 & 92.39 & 93.53 & 84.91 \\
Llama-3-70B-8192 & 96.36 & 97.83 & 91.55 \\
\midrule
Llama-4-Scout-17B-16e-instruct & 96.45 & 97.21 & 86.69 \\
Llama-4-Maverick-17B-128e-instruct & 96.96 & \textbf{98.15} & 94.35 \\
\midrule
GPT-4o-mini & 96.95 & 96.79 & 88.42 \\
GPT-4o & \textbf{99.64} & 97.36 & \textbf{99.16} \\
\midrule
DeepSeek-R1-Distill-Llama-70B & 91.38 & 90.26 & 88.94 \\
Qwen-QWQ-32B & 91.73 & 90.83 & 90.08 \\
\bottomrule
\end{tabular}
\end{table*}

\begin{table*}[!htbp]
\caption{LLM Factor Utilization Recall ($Rec_U$, \% Across Tests 1 \& 2)}
\label{tab:recall_results}
\centering
\footnotesize
\setlength{\tabcolsep}{6pt} 
\begin{tabular}{lcc} 
\toprule
\textbf{Model} & \textbf{Test 1 (Arguable)} & \textbf{Test 2 (Reordered)} \\ 
\midrule
Llama-3-8B-8192 & 64.63 & 61.42 \\
Llama-3-70B-8192 & 77.51 & 74.62 \\
\midrule
Llama-4-Scout-17B-16e-instruct & 69.47 & 63.77 \\
Llama-4-Maverick-17B-128e-instruct & 53.95 & 51.40 \\
\midrule
GPT-4o-mini & 42.47 & 49.55 \\
GPT-4o & \textbf{85.22} & \textbf{76.61} \\
\midrule
DeepSeek-R1-Distill-Llama-70B & 72.61 & 69.58 \\
Qwen-QWQ-32B & 61.40 & 52.51 \\
\bottomrule
\end{tabular}
\end{table*}

\begin{table*}[!htbp]
\caption{LLM Abstention Ratio ($Ratio_{Abstain}$, \% for Test 3)}
\label{tab:abstention_ratio_results}
\centering
\footnotesize
\setlength{\tabcolsep}{6pt} 
\begin{tabular}{lc}
\toprule
\textbf{Model} & \textbf{Abstention Ratio (Test 3)} \\
\midrule
Llama-3-8B-8192 & 0.00 \\
Llama-3-70B-8192 & 3.33 \\
\midrule
Llama-4-Scout-17B-16e-instruct & 0.00 \\
Llama-4-Maverick-17B-128e-instruct & 50.00 \\
\midrule
GPT-4o-mini & 0.00 \\
GPT-4o & \textbf{86.67} \\
\midrule
DeepSeek-R1-Distill-Llama-70B & 23.33 \\
Qwen-QWQ-32B & 56.67 \\
\bottomrule
\end{tabular}
\end{table*}

\subsection{Hallucination Accuracy Results}
As shown in Table \ref{tab:accuracy_results}, the Hallucination Accuracy ($Acc_H$) is generally very high for Tests 1 and 2 across most models. These tests involve generating arguments based on provided `Arguable' scenarios, either in a standard format (Test 1) or with swapped precedent roles (Test 2). GPT-4o achieves near-perfect accuracy (>99\% for Test 1, >97\% for Test 2), indicating exceptional faithfulness in citing only those factors genuinely shared between the relevant cases in these scenarios. Other models, including GPT-4o-mini, Llama-3-70B, and the Llama-4 variants, also demonstrate high accuracy, typically above 96\% on these tests. This suggests that when instructed to generate arguments based on factor representations where supporting shared factors exist, current leading LLMs are capable of doing so with minimal hallucination according to our metric. Smaller or specialized models like DeepSeek, Qwen, and Llama-3-8B show slightly lower but still respectable accuracy, generally above 90\%.

For Test 3 (Non-arguable), where no shared factors exist and models were instructed to abstain, the $Acc_H$ remains surprisingly high for several models, particularly GPT-4o (99.16\%) and Llama-4-Maverick (94.35\%). This high accuracy, however, must be interpreted cautiously alongside the primary goal of abstention and the recall results (Section 5.3).

\subsection{Factor Utilization Recall Results}
Factor Utilization Recall ($Rec_U$), presented in Table \ref{tab:recall_results}, measures the completeness of the arguments by quantifying the proportion of available supporting factors that were correctly identified and used by the LLM for Tests 1 and 2. The results reveal a more varied picture than accuracy. For Tests 1 and 2, GPT-4o again leads, utilizing over 85\% of available factors in the standard test (Test 1) and over 76\% in the swapped-role test (Test 2). Llama-3-70B also performs strongly, achieving recall scores above 74\% for both tests. Other models exhibit a wider range of performance. For instance, GPT-4o-mini shows significantly lower recall (around 42-50\%), suggesting it generates arguments that are factually accurate (high $Acc_H$) but lack comprehensiveness. The Llama-4 variants, DeepSeek, Qwen, and Llama-3-8B fall between these extremes, with recall generally ranging from 50\% to 70\% on the arguable tests. This indicates that while models can avoid making up facts, they often fail to incorporate the full set of relevant facts available in the input materials into their generated arguments.

\subsection{Abstention Test Performance}
Test 3 was designed specifically to test the models' ability to follow the instruction to abstain when faced with `Non-arguable' scenarios lacking shared factors. The primary metric for this test is the Abstention Ratio ($Ratio_{Abstain}$), presented in Table \ref{tab:abstention_ratio_results}. We also consider Hallucination Accuracy ($Acc_H$) from Table \ref{tab:accuracy_results} for Test 3 to understand the nature of arguments generated when models failed to abstain.

As shown in Table \ref{tab:abstention_ratio_results}, the ability to correctly abstain varies significantly across models. GPT-4o achieved the highest abstention ratio (86.67\%), successfully following the instruction in the majority of non-arguable cases. Qwen-QWQ-32B (56.67\%) and Llama-4-Maverick (50.00\%) also demonstrated some capability to abstain. However, several models, including Llama-3-8B, Llama-4-Scout, and GPT-4o-mini, had an abstention ratio of 0.00\%, indicating they failed to abstain in any of the test instances. Llama-3-70B also performed poorly with a very low abstention ratio (3.33\%).

For models that failed to abstain and instead generated spurious arguments, their Hallucination Accuracy ($Acc_H$) on Test 3 (Table \ref{tab:accuracy_results}) is informative. For example, GPT-4o, even when it rarely failed to abstain, maintained very high $Acc_H$ (99.16\%), meaning its spurious arguments were largely free of hallucinated factors not in the input cases. Llama-4-Maverick also showed high $Acc_H$ (94.35\%) in such instances. This suggests that their failure was primarily in not following the abstention instruction, rather than fabricating factors. Other models that failed to abstain also generally maintained relatively high $Acc_H$ (mostly above 84\%), indicating that the spurious arguments, while incorrect, were mostly based on factors present in the input cases rather than completely fabricated information.

Overall, this critical test of instruction following reveals a significant weakness in most LLMs. The inability to reliably recognize task impossibility and adhere to negative constraints is a major concern for their deployment in sensitive applications. Even models that performed well on argument generation (Tests 1 \& 2) struggled significantly with abstention.

\subsection{Comparative Analysis}
Across Tests 1 and 2 (argument generation), GPT-4o demonstrates the strongest performance, achieving the highest Hallucination Accuracy and leading significantly in Factor Utilization Recall, suggesting its arguments are both faithful and comprehensive. Llama-3-70B generally ranks second, showing strong accuracy and good recall. Llama-4-Maverick also performs well in terms of accuracy on these tests, though its recall is moderate.

The performance gap between the top models (GPT-4o, Llama-3-70B) and others is more pronounced in Factor Utilization Recall than in Hallucination Accuracy for Tests 1 and 2. Models like GPT-4o-mini, Llama-3-8B, and Qwen often achieve reasonable accuracy but struggle with recall, producing less complete arguments. The Llama-4 variants and DeepSeek fall in the mid-range for both metrics on these arguable tests.

Test 3 (Abstention) reveals a critical dimension of model capability. GPT-4o stands out with the highest Abstention Ratio (Table \ref{tab:abstention_ratio_results}), indicating a superior ability to follow instructions to abstain. Qwen-QWQ-32B and Llama-4-Maverick show moderate success in abstention, while several models, including some high performers on Tests 1 \& 2 like Llama-3-70B, almost completely failed to abstain. This highlights that strong performance on generative tasks does not necessarily translate to robust instruction following for negative constraints. The failure to abstain, even when explicitly instructed, is a significant concern. The failure modes on Test 3 varied, but very few models performed the task as intended by consistently and correctly abstaining.

\section{Error Analysis}

To gain deeper insights into the quantitative results and understand the nature of the errors identified by our automated metrics, we conducted an error analysis. We selected LLM outputs for manual review primarily based on instances where the automated metrics indicated significant deviations from desired performance. This included cases with lower Hallucination Accuracy ($Acc_H < 95\%$), notably low Factor Utilization Recall ($Rec_U$), outputs from models exhibiting generally weaker performance on a specific test, and, critically, all instances where models failed to produce the correct abstention output on Test 3. The analysis involved a  manual review of the selected LLM-generated argument texts. Each generated argument was compared against the ground-truth factors provided in the corresponding input case triple (Current Case, TSC1, TSC2) and the specific instructions given in the core prompt (including the 3-ply structure requirements and the abstention rule). During the review, observed errors were categorized into distinct types related to hallucination, incomplete factor utilization, and failures in following instructions, particularly regarding the abstention task.

\textbf{Hallucination Errors (Primarily Tests 1 \& 2):} Although quantitative results showed high $Acc_H$, qualitative analysis identified infrequent instances, primarily in lower-performing models.
\begin{itemize}[noitemsep, topsep=0pt]
    \item \textit{Factor Misattribution:} Citing a factor as present in one case (e.g., the Current Case) when it actually belonged to a different case in the input triple (e.g., TSC1).
\end{itemize}

\begin{custombox}{Factor Misattribution: GPT-4o}
    \label{box:hallucination_errors} 
    ``Plaintiff's Argument: ... Factors F3 ... F21 Knew-info-confidential (P), F23 Waiver-of-confidentiality (D), and F25 Info-reverse-engineered (D) were present in both the input case and TSC1 ...''
\end{custombox}
For example, in one instance of a GPT-4o output (Box~\ref{box:hallucination_errors}), factor F21 was incorrectly attributed to TSC1 by the model, as it was not present in the ground-truth factors for TSC1.


\textbf{Incomplete Factor Utilization Errors (Tests 1 \& 2):} This was a more common issue across models, reflected in the $Rec_U$ scores.
\begin{itemize}[noitemsep, topsep=0pt]
    \item \textit{Omission of Shared Factors:} Failing to identify or mention relevant factors that were shared between the Current Case and the precedent being cited (TSC1 for Plaintiff's first ply, TSC2 for Defendant's ply).
    \item \textit{Omission of Distinguishing Factors:} Failing to identify or mention factors that differentiate the Current Case from the precedent being discussed, particularly when the task required distinguishing (Defendant's ply distinguishing TSC1, Plaintiff's rebuttal distinguishing TSC2).
\end{itemize}

\begin{custombox}{Omission of Shared Factors \& Omission of Distinguishing Factors: Llama-3-8B-8192}
    \label{box:incomplete_factor_utilization_errors} 
    ``Plaintiff's Argument: ... Factors F4 Agreed-not-to-disclose (P) and F6 Security-measures (P) were present in both the input case and TSC1 ...'' \\
    ``Plaintiff's Rebuttal: ... In TSC2, the additional factors ... F25 Info-reverse-engineered (D), F27 Disclosure-in-public-forum (D) were present and are not present in input case.''
\end{custombox}
For instance, with Llama-3-8B-8192 (Box~\ref{box:incomplete_factor_utilization_errors}), the model failed to mention that F7 Brought-tools (P) was a shared pro-plaintiff factor in both the current case and TSC1 during the Plaintiff's Argument. Additionally, in the Plaintiff's Rebuttal, the model did not point out that F12 Outsider-disclosures-restricted (P) was a distinguishing factor present in the current case but not in TSC2.

\textbf{Instruction Following / Abstention Errors (Test 3):} This was the most significant failure mode observed across nearly all models.
\begin{itemize}[noitemsep, topsep=0pt]
    \item \textit{Failure to Abstain:} The most common error was generating a spurious 3-ply argument structure despite the lack of common factors and the explicit instruction to output a specific abstention phrase. We observe that even in instances where the model does not successfully abstain, the resulting legal arguments do not exhibit factual misattribution or hallucination.
    \item \textit{Incorrect Abstention Phrase:} In rare cases where a model attempted to signal impossibility, it failed to use the exact required phrase (``No common factor between the current case and the TSC1/TSC2''), using a variation instead.
    \item \textit{Spurious Reasoning within Failed Abstention:} When models incorrectly generated arguments, the reasoning was inherently flawed. This often involved discussing factors from the cases as if they provided a basis for comparison, even though no common factors existed, or fabricating connections.
\end{itemize}

\begin{custombox}{Failure to Abstain: GPT-4o}
    \label{box:failure_to_abstain} 
    ``Plaintiff's Argument: ... F22 Invasive-techniques (P), F26 Deception (P) were present in the input case and support the Plaintiff's position ...''
\end{custombox}
This output (Box~\ref{box:failure_to_abstain}) illustrates GPT-4o failing to follow the instruction to abstain when presented with a non-arguable scenario in Test 3. Instead of outputting the specified abstention phrase, the model proceeded to generate an argument. Notably, while the generation of any argument was incorrect in this context, the factors cited in this whole output were present in the input case material, indicating a failure in instruction adherence rather than a hallucination errors.

\begin{custombox}{Spurious Reasoning within Failed Abstention: Deepseek-R1-Distill-Llama-70B}
    \label{box:spurious_reasoning_within_failed_abstention} 
    ``<think> ... The first step is to check if there are common factors between the input current case and either TSC1 or TSC2. If there are none, I should stop ... <\textbackslash think>'' \\ ``Plaintiff's Argument: Factors F6 Security-measures (P) and F21 Knew-info-confidential (P) were present in both the input case and TSC1 ...''
\end{custombox}
As shown in the example (Box~\ref{box:spurious_reasoning_within_failed_abstention}), even when a model's reasoning trace explicitly mentioned the condition for stopping, it might still proceed to generate a spurious argument, failing to follow the critical abstention instruction. Additionally, the model claimed that there were common factors between the input case and TSC1.

This qualitative analysis complements the quantitative metrics, highlighting that even with high accuracy, completeness remains a challenge, and adherence to negative constraints like abstention is a critical weakness for current LLMs in this legal argument generation context.

\section{Conclusion}

This paper introduced and applied an automated pipeline to evaluate the performance of eight LLMs on generating 3-ply, factor-based legal arguments, focusing specifically on faithfulness, completeness, and the ability to follow abstention instructions. Our evaluation, guided by three research questions, yielded the following conclusions:

Regarding \textbf{RQ1} (hallucination and incomplete factor utilization), our results show that while most evaluated LLMs exhibit high Hallucination Accuracy ($Acc_H > 90-95\%$ in Tests 1 \& 2), indicating they generally avoid citing non-existent factors when generating arguments in viable scenarios, they struggle with completeness. Factor Utilization Recall ($Rec_U$) varied significantly (from $\approx$40\% to $\approx$85\% in Tests 1 \& 2), demonstrating that LLMs often omit relevant, available factors from the input cases, leading to potentially superficial arguments.

Concerning \textbf{RQ2} (adherence to abstention instructions), the evaluation revealed a critical weakness across almost all models. When presented with non-arguable scenarios (Test 3) and explicitly instructed to abstain, most models failed to follow this directive, as measured by our Abstention Ratio (Table \ref{tab:abstention_ratio_results}). Instead of abstaining, they generated spurious arguments. Only a few models showed a significant ability to abstain, with GPT-4o performing best, yet still not perfectly. This highlights a fundamental inability in most current LLMs to reliably recognize task impossibility and follow negative constraints.

Addressing \textbf{RQ3} (effectiveness of automated metrics), the proposed metrics ($Acc_H$, $Rec_U$, and $Ratio_{Abstain}$) successfully quantified distinct error types. $Acc_H$ effectively measured faithfulness (absence of hallucination). $Rec_U$ captured incomplete factor utilization in argument generation tasks (Tests 1 \& 2). $Ratio_{Abstain}$ directly measured the critical capability of adherence to abstention instructions in Test 3. Together, these metrics clearly revealed performance variations across different LLMs and task complexities, demonstrating the pipeline's utility in diagnosing specific weaknesses.

In summary, while LLMs show promise in generating factually grounded components of legal arguments based on structured inputs, significant improvements are needed in ensuring comprehensive reasoning (completeness) and, most crucially, in robust instruction following, particularly regarding negative constraints and the ability to abstain appropriately. These deficiencies must be addressed before LLMs can be reliably deployed for substantive legal argumentation tasks.

\section{Limitations and Future Work}

This study is subject to several limitations. The evaluation utilizes synthetic, factor-represented cases, simplifying the nuances of real-world legal texts and reasoning. The accuracy of our automated metrics inherently depends on the performance of the external LLM used for factor extraction, introducing a potential layer of error. The specific operationalization of our metrics, particularly for the abstention test, could be further refined. Furthermore, the findings are based on a specific dataset size, prompt structure, and set of LLMs, potentially limiting the generalizability of the precise quantitative results, although the qualitative trends are likely indicative.

Future research should aim to address these limitations. Evaluating performance on larger, more diverse datasets, including those derived from real-world legal documents (which would necessitate robust factor extraction from text as a preliminary step \cite{gray2023automatic}), is a crucial next step. Further validation and refinement of the automated metrics, potentially including comparisons with human expert judgments on argument quality beyond factor usage, would strengthen the evaluation pipeline. Investigating the underlying reasons for the observed deficiencies in recall and abstention through model interpretability or targeted probing could inform the development of more reliable models. Finally, exploring novel prompting strategies, fine-tuning approaches, or architectural modifications specifically designed to enhance completeness and robust instruction adherence in legal argument generation remains a vital avenue for future work.


  

\bibliography{Reference}

\begin{thebibliography}{31}
\expandafter\ifx\csname natexlab\endcsname\relax\def\natexlab#1{#1}\fi
\providecommand{\url}[1]{\texttt{#1}}
\providecommand{\href}[2]{#2}
\providecommand{\path}[1]{#1}
\providecommand{\DOIprefix}{doi:}
\providecommand{\ArXivprefix}{arXiv:}
\providecommand{\URLprefix}{URL: }
\providecommand{\Pubmedprefix}{pmid:}
\providecommand{\doi}[1]{\href{http://dx.doi.org/#1}{\path{#1}}}
\providecommand{\Pubmed}[1]{\href{pmid:#1}{\path{#1}}}
\providecommand{\bibinfo}[2]{#2}
\ifx\xfnm\relax \def\xfnm[#1]{\unskip,\space#1}\fi
\bibitem[{Ariai and Demartini(2024)}]{ariai2024natural}
\bibinfo{author}{F.~Ariai}, \bibinfo{author}{G.~Demartini},
\newblock \bibinfo{title}{Natural language processing for the legal domain: A survey of tasks, datasets, models, and challenges},
\newblock \bibinfo{journal}{arXiv preprint arXiv:2410.21306}  (\bibinfo{year}{2024}).
\bibitem[{Katz et~al.(2023)Katz, Hartung, Gerlach, Jana, and Bommarito~II}]{katz2023natural}
\bibinfo{author}{D.~M. Katz}, \bibinfo{author}{D.~Hartung}, \bibinfo{author}{L.~Gerlach}, \bibinfo{author}{A.~Jana}, \bibinfo{author}{M.~J. Bommarito~II},
\newblock \bibinfo{title}{Natural language processing in the legal domain},
\newblock \bibinfo{journal}{arXiv preprint arXiv:2302.12039}  (\bibinfo{year}{2023}).
\bibitem[{Gray et~al.(2025)Gray, Zhang, and Ashley}]{gray2025generating}
\bibinfo{author}{M.~A. Gray}, \bibinfo{author}{L.~Zhang}, \bibinfo{author}{K.~D. Ashley},
\newblock \bibinfo{title}{Generating case-based legal arguments with llms},
\newblock in: \bibinfo{booktitle}{Proceedings of the 4th ACM Computers and Law Symposium}, \bibinfo{year}{2025}.
\bibitem[{Bommasani et~al.(2021)Bommasani, Hudson, Adeli, Altman, Arora, von Arx, Bernstein, Bohg, Bosselut, Brunskill et~al.}]{bommasani2021opportunities}
\bibinfo{author}{R.~Bommasani}, \bibinfo{author}{D.~A. Hudson}, \bibinfo{author}{E.~Adeli}, \bibinfo{author}{R.~Altman}, \bibinfo{author}{S.~Arora}, \bibinfo{author}{S.~von Arx}, \bibinfo{author}{M.~S. Bernstein}, \bibinfo{author}{J.~Bohg}, \bibinfo{author}{A.~Bosselut}, \bibinfo{author}{E.~Brunskill}, et~al.,
\newblock \bibinfo{title}{On the opportunities and risks of foundation models},
\newblock \bibinfo{journal}{arXiv preprint arXiv:2108.07258}  (\bibinfo{year}{2021}).
\bibitem[{Ji et~al.(2023)Ji, Lee, Frieske, Yu, Su, Xu, Ishii, Bang, Madotto, and Fung}]{ji2023survey}
\bibinfo{author}{Z.~Ji}, \bibinfo{author}{N.~Lee}, \bibinfo{author}{R.~Frieske}, \bibinfo{author}{T.~Yu}, \bibinfo{author}{D.~Su}, \bibinfo{author}{Y.~Xu}, \bibinfo{author}{E.~Ishii}, \bibinfo{author}{Y.~J. Bang}, \bibinfo{author}{A.~Madotto}, \bibinfo{author}{P.~Fung},
\newblock \bibinfo{title}{Survey of hallucination in natural language generation},
\newblock \bibinfo{journal}{ACM Computing Surveys} \bibinfo{volume}{55} (\bibinfo{year}{2023}) \bibinfo{pages}{1--38}.
\bibitem[{de~la Osa and Remolina(2024)}]{de2024artificial}
\bibinfo{author}{D.~U.~S. de~la Osa}, \bibinfo{author}{N.~Remolina},
\newblock \bibinfo{title}{Artificial intelligence at the bench: Legal and ethical challenges of informing—or misinforming—judicial decision-making through generative ai},
\newblock \bibinfo{journal}{Data \& Policy} \bibinfo{volume}{6} (\bibinfo{year}{2024}) \bibinfo{pages}{e59}.
\bibitem[{Avery et~al.(2023)Avery, Abril, and del Riego}]{avery2023chatgpt}
\bibinfo{author}{J.~J. Avery}, \bibinfo{author}{P.~S. Abril}, \bibinfo{author}{A.~del Riego},
\newblock \bibinfo{title}{Chatgpt, esq.: Recasting unauthorized practice of law in the era of generative ai},
\newblock \bibinfo{journal}{Yale JL \& Tech.} \bibinfo{volume}{26} (\bibinfo{year}{2023}) \bibinfo{pages}{64}.
\bibitem[{Guha et~al.(2024)Guha, Nyarko, Ho, R{\'e}, Chilton, Chohlas-Wood, Peters, Waldon, Rockmore, Zambrano et~al.}]{guha2024legalbench}
\bibinfo{author}{N.~Guha}, \bibinfo{author}{J.~Nyarko}, \bibinfo{author}{D.~Ho}, \bibinfo{author}{C.~R{\'e}}, \bibinfo{author}{A.~Chilton}, \bibinfo{author}{A.~Chohlas-Wood}, \bibinfo{author}{A.~Peters}, \bibinfo{author}{B.~Waldon}, \bibinfo{author}{D.~Rockmore}, \bibinfo{author}{D.~Zambrano}, et~al.,
\newblock \bibinfo{title}{Legalbench: A collaboratively built benchmark for measuring legal reasoning in large language models},
\newblock \bibinfo{journal}{Advances in Neural Information Processing Systems} \bibinfo{volume}{36} (\bibinfo{year}{2024}).
\bibitem[{Li et~al.(2024)Li, Chen, Ai, Wu, Zhang, and Liu}]{li2024lexeval}
\bibinfo{author}{H.~Li}, \bibinfo{author}{Y.~Chen}, \bibinfo{author}{Q.~Ai}, \bibinfo{author}{Y.~Wu}, \bibinfo{author}{R.~Zhang}, \bibinfo{author}{Y.~Liu},
\newblock \bibinfo{title}{Lexeval: A comprehensive chinese legal benchmark for evaluating large language models},
\newblock \bibinfo{journal}{arXiv preprint arXiv:2409.20288}  (\bibinfo{year}{2024}).
\bibitem[{Fei et~al.(2023)Fei, Shen, Zhu, Zhou, Han, Zhang, Chen, Shen, and Ge}]{fei2023lawbench}
\bibinfo{author}{Z.~Fei}, \bibinfo{author}{X.~Shen}, \bibinfo{author}{D.~Zhu}, \bibinfo{author}{F.~Zhou}, \bibinfo{author}{Z.~Han}, \bibinfo{author}{S.~Zhang}, \bibinfo{author}{K.~Chen}, \bibinfo{author}{Z.~Shen}, \bibinfo{author}{J.~Ge},
\newblock \bibinfo{title}{Lawbench: Benchmarking legal knowledge of large language models},
\newblock \bibinfo{journal}{arXiv preprint arXiv:2309.16289}  (\bibinfo{year}{2023}).
\bibitem[{Touvron et~al.(2023)Touvron, Lavril, Izacard, Martinet, Lachaux, Lacroix, Rozi{\`e}re, Goyal, Hambro, Azhar et~al.}]{touvron2023llama}
\bibinfo{author}{H.~Touvron}, \bibinfo{author}{T.~Lavril}, \bibinfo{author}{G.~Izacard}, \bibinfo{author}{X.~Martinet}, \bibinfo{author}{M.-A. Lachaux}, \bibinfo{author}{T.~Lacroix}, \bibinfo{author}{B.~Rozi{\`e}re}, \bibinfo{author}{N.~Goyal}, \bibinfo{author}{E.~Hambro}, \bibinfo{author}{F.~Azhar}, et~al.,
\newblock \bibinfo{title}{Llama: Open and efficient foundation language models},
\newblock \bibinfo{journal}{arXiv preprint arXiv:2302.13971}  (\bibinfo{year}{2023}).
\bibitem[{Achiam et~al.(2023)Achiam, Adler, Agarwal, Ahmad, Akkaya, Aleman, Almeida, Altenschmidt, Altman, Anadkat et~al.}]{achiam2023gpt}
\bibinfo{author}{J.~Achiam}, \bibinfo{author}{S.~Adler}, \bibinfo{author}{S.~Agarwal}, \bibinfo{author}{L.~Ahmad}, \bibinfo{author}{I.~Akkaya}, \bibinfo{author}{F.~L. Aleman}, \bibinfo{author}{D.~Almeida}, \bibinfo{author}{J.~Altenschmidt}, \bibinfo{author}{S.~Altman}, \bibinfo{author}{S.~Anadkat}, et~al.,
\newblock \bibinfo{title}{Gpt-4 technical report},
\newblock \bibinfo{journal}{arXiv preprint arXiv:2303.08774}  (\bibinfo{year}{2023}).
\bibitem[{Ashley(1990)}]{ashley1990}
\bibinfo{author}{K.~D. Ashley}, \bibinfo{title}{Modeling Legal Argument: Reasoning with Cases and Hypotheticals}, \bibinfo{publisher}{The MIT Press}, \bibinfo{address}{Cambridge, MA}, \bibinfo{year}{1990}.
\bibitem[{Aleven and Ashley(1997)}]{aleven1997teaching}
\bibinfo{author}{V.~Aleven}, \bibinfo{author}{K.~D. Ashley},
\newblock \bibinfo{title}{Teaching case-based argumentation through a model and examples: Empirical evaluation of an intelligent learning environment},
\newblock in: \bibinfo{booktitle}{Artificial intelligence in education}, volume~\bibinfo{volume}{39}, \bibinfo{organization}{Citeseer}, \bibinfo{year}{1997}, pp. \bibinfo{pages}{87--94}.
\bibitem[{Rissland and Skalak(1991)}]{rissland1991cabaret}
\bibinfo{author}{E.~L. Rissland}, \bibinfo{author}{D.~B. Skalak},
\newblock \bibinfo{title}{Cabaret:rule interpretation in a hybrid architecture},
\newblock \bibinfo{journal}{International Journal of Man-machine Studies} \bibinfo{volume}{34} (\bibinfo{year}{1991}) \bibinfo{pages}{839–887}.
\bibitem[{Br{\"u}ninghaus and Ashley(2003)}]{bruninghaus2003predicting}
\bibinfo{author}{S.~Br{\"u}ninghaus}, \bibinfo{author}{K.~D. Ashley},
\newblock \bibinfo{title}{Predicting outcomes of case based legal arguments},
\newblock in: \bibinfo{booktitle}{Proceedings of the 9th International conference on Artificial Intelligence and Law}, \bibinfo{organization}{ACM}, \bibinfo{year}{2003}, pp. \bibinfo{pages}{233--242}.
\bibitem[{Grabmair(2017)}]{grabmair2017predicting}
\bibinfo{author}{M.~Grabmair},
\newblock \bibinfo{title}{Predicting trade secret case outcomes using argument schemes and learned quantitative value effect tradeoffs},
\newblock in: \bibinfo{booktitle}{Proceedings of the 16th ICAIL}, \bibinfo{year}{2017}, pp. \bibinfo{pages}{89--98}.
\bibitem[{Horty(2021)}]{horty2021modifying}
\bibinfo{author}{J.~F. Horty},
\newblock \bibinfo{title}{Modifying precedential constraint},
\newblock \bibinfo{journal}{Journal of Artificial Intelligence Law} \bibinfo{volume}{30} (\bibinfo{year}{2021}) \bibinfo{pages}{1--24}.
\bibitem[{Gray et~al.(2024)Gray, Savelka, Oliver, and Ashley}]{gray2024empirical}
\bibinfo{author}{M.~A. Gray}, \bibinfo{author}{J.~Savelka}, \bibinfo{author}{W.~M. Oliver}, \bibinfo{author}{K.~D. Ashley},
\newblock \bibinfo{title}{Empirical legal analysis simplified: reducing complexity through automatic identification and evaluation of legally relevant factors},
\newblock \bibinfo{journal}{Philosophical Transactions of the Royal Society A} \bibinfo{volume}{382} (\bibinfo{year}{2024}) \bibinfo{pages}{20230155}.
\bibitem[{Savelka and Ashley(2023)}]{savelka2023unreasonable}
\bibinfo{author}{J.~Savelka}, \bibinfo{author}{K.~D. Ashley},
\newblock \bibinfo{title}{The unreasonable effectiveness of large language models in zero-shot semantic annotation of legal texts},
\newblock \bibinfo{journal}{Frontiers in Artificial Intelligence} \bibinfo{volume}{6} (\bibinfo{year}{2023}) \bibinfo{pages}{1279794}.
\bibitem[{Yadkori et~al.(2024)Yadkori, Kuzborskij, Gy{\"o}rgy, and Szepesv{\'a}ri}]{yadkori2024believe}
\bibinfo{author}{Y.~A. Yadkori}, \bibinfo{author}{I.~Kuzborskij}, \bibinfo{author}{A.~Gy{\"o}rgy}, \bibinfo{author}{C.~Szepesv{\'a}ri},
\newblock \bibinfo{title}{To believe or not to believe your llm},
\newblock \bibinfo{journal}{arXiv preprint arXiv:2406.02543}  (\bibinfo{year}{2024}).
\bibitem[{Zhang et~al.(2023)Zhang, Li, Cui, Cai, Liu, Fu, Huang, Zhao, Zhang, Chen et~al.}]{zhang2023siren}
\bibinfo{author}{Y.~Zhang}, \bibinfo{author}{Y.~Li}, \bibinfo{author}{L.~Cui}, \bibinfo{author}{D.~Cai}, \bibinfo{author}{L.~Liu}, \bibinfo{author}{T.~Fu}, \bibinfo{author}{X.~Huang}, \bibinfo{author}{E.~Zhao}, \bibinfo{author}{Y.~Zhang}, \bibinfo{author}{Y.~Chen}, et~al.,
\newblock \bibinfo{title}{Siren's song in the ai ocean: a survey on hallucination in large language models},
\newblock \bibinfo{journal}{arXiv preprint arXiv:2309.01219}  (\bibinfo{year}{2023}).
\bibitem[{Dhuliawala et~al.(2023)Dhuliawala, Komeili, Xu, Raileanu, Li, Celikyilmaz, and Weston}]{dhuliawala2023chain}
\bibinfo{author}{S.~Dhuliawala}, \bibinfo{author}{M.~Komeili}, \bibinfo{author}{J.~Xu}, \bibinfo{author}{R.~Raileanu}, \bibinfo{author}{X.~Li}, \bibinfo{author}{A.~Celikyilmaz}, \bibinfo{author}{J.~Weston},
\newblock \bibinfo{title}{Chain-of-verification reduces hallucination in large language models},
\newblock \bibinfo{journal}{arXiv preprint arXiv:2309.11495}  (\bibinfo{year}{2023}).
\bibitem[{Vu et~al.(2023)Vu, Iyyer, Wang, Constant, Wei, Wei, Tar, Sung, Zhou, Le et~al.}]{vu2023freshllms}
\bibinfo{author}{T.~Vu}, \bibinfo{author}{M.~Iyyer}, \bibinfo{author}{X.~Wang}, \bibinfo{author}{N.~Constant}, \bibinfo{author}{J.~Wei}, \bibinfo{author}{J.~Wei}, \bibinfo{author}{C.~Tar}, \bibinfo{author}{Y.-H. Sung}, \bibinfo{author}{D.~Zhou}, \bibinfo{author}{Q.~Le}, et~al.,
\newblock \bibinfo{title}{Freshllms: Refreshing large language models with search engine augmentation},
\newblock \bibinfo{journal}{arXiv preprint arXiv:2310.03214}  (\bibinfo{year}{2023}).
\bibitem[{Lin(2004)}]{lin2004rouge}
\bibinfo{author}{C.-Y. Lin},
\newblock \bibinfo{title}{Rouge: A package for automatic evaluation of summaries},
\newblock in: \bibinfo{booktitle}{Text summarization branches out}, \bibinfo{year}{2004}, pp. \bibinfo{pages}{74--81}.
\bibitem[{Papineni et~al.(2002)Papineni, Roukos, Ward, and Zhu}]{papineni2002bleu}
\bibinfo{author}{K.~Papineni}, \bibinfo{author}{S.~Roukos}, \bibinfo{author}{T.~Ward}, \bibinfo{author}{W.-J. Zhu},
\newblock \bibinfo{title}{Bleu: a method for automatic evaluation of machine translation},
\newblock in: \bibinfo{booktitle}{Proceedings of the 40th annual meeting of the Association for Computational Linguistics}, \bibinfo{year}{2002}, pp. \bibinfo{pages}{311--318}.
\bibitem[{Zhang et~al.(2019)Zhang, Kishore, Wu, Weinberger, and Artzi}]{zhang2019bertscore}
\bibinfo{author}{T.~Zhang}, \bibinfo{author}{V.~Kishore}, \bibinfo{author}{F.~Wu}, \bibinfo{author}{K.~Q. Weinberger}, \bibinfo{author}{Y.~Artzi},
\newblock \bibinfo{title}{Bertscore: Evaluating text generation with bert},
\newblock \bibinfo{journal}{arXiv preprint arXiv:1904.09675}  (\bibinfo{year}{2019}).
\bibitem[{Wen et~al.(2024)Wen, Yao, Feng, Xu, Tsvetkov, Howe, and Wang}]{wen2024know}
\bibinfo{author}{B.~Wen}, \bibinfo{author}{J.~Yao}, \bibinfo{author}{S.~Feng}, \bibinfo{author}{C.~Xu}, \bibinfo{author}{Y.~Tsvetkov}, \bibinfo{author}{B.~Howe}, \bibinfo{author}{L.~L. Wang},
\newblock \bibinfo{title}{Know your limits: A survey of abstention in large language models},
\newblock \bibinfo{journal}{arXiv preprint arXiv:2407.18418}  (\bibinfo{year}{2024}).
\bibitem[{Feng et~al.(2024)Feng, Shi, Wang, Ding, Balachandran, and Tsvetkov}]{feng2024don}
\bibinfo{author}{S.~Feng}, \bibinfo{author}{W.~Shi}, \bibinfo{author}{Y.~Wang}, \bibinfo{author}{W.~Ding}, \bibinfo{author}{V.~Balachandran}, \bibinfo{author}{Y.~Tsvetkov},
\newblock \bibinfo{title}{Don't hallucinate, abstain: Identifying llm knowledge gaps via multi-llm collaboration},
\newblock \bibinfo{journal}{arXiv preprint arXiv:2402.00367}  (\bibinfo{year}{2024}).
\bibitem[{Ashley(2017)}]{ashley2017artificial}
\bibinfo{author}{K.~D. Ashley}, \bibinfo{title}{Artificial intelligence and legal analytics: new tools for law practice in the digital age}, \bibinfo{publisher}{Cambridge University Press}, \bibinfo{year}{2017}.
\bibitem[{Gray et~al.(2023)Gray, Savelka, Oliver, and Ashley}]{gray2023automatic}
\bibinfo{author}{M.~Gray}, \bibinfo{author}{J.~Savelka}, \bibinfo{author}{W.~Oliver}, \bibinfo{author}{K.~Ashley},
\newblock \bibinfo{title}{Automatic identification and empirical analysis of legally relevant factors},
\newblock in: \bibinfo{booktitle}{Proceedings of the Nineteenth International Conference on Artificial Intelligence and Law}, \bibinfo{year}{2023}, pp. \bibinfo{pages}{101--110}.

\end{thebibliography}

\appendix
\section{Prompt Structure for 3-Ply Argument Generation}
\label{app:example_prompt}

The following shows the structure of the prompt provided to the LLMs for the 3-ply argument generation task.

\begin{custombox}{Example Prompt}
    \label{prompt:ts_misappropriation_appendix} 
    \textbf{TASK}\\
    In this task, we will formulate legal arguments based on trade secret misappropriation claims using a structured approach. Follow the steps outlined below for consistency and clarity.

    \textbf{Legal Problem Context}\\
    In this problem, we aim to develop arguments using factors critical to trade secret misappropriation claims. Typically, the Plaintiff alleges that the Defendant has misappropriated their trade secret. For instance, Kentucky Fried Chicken (KFC) could claim misappropriation if an employee disclosed their secret recipe, which is a blend of herbs and spices, by publishing it in a cookbook.

    Factors may support either the Plaintiff (P) or the Defendant (D). The Plaintiff might emphasize measures they took to protect the recipe, while the Defendant could argue that the recipe was already disclosed to outsiders. Based on the factors provided, construct a three-part argument as detailed below.

    \textbf{Instructions}
    \begin{enumerate}
        \item \textbf{IMPORTANT:} If there is no common factor between the current case and the TSC1/TSC2, you need to say "No common factor between the input current case and the TSC1/TSC2" and stop generating any argument.
        \item \textbf{Construct a 3-Ply Argument:}
        \begin{enumerate}
            \item \textbf{Plaintiff's Argument:} Present an argument in favor of the Plaintiff's position by i. citing a relevant Trade Secret Case (TSC1/TSC2) with a similar favorable outcome; ii. Highlighting shared factors between the input current case and the TSC1/TSC2.
            \item \textbf{Defendant's Counterargument:} Refute the Plaintiff's position by i. Distinguishing the cited TSC1/TSC2 based on differing factors; ii. Citing a counterexample (a TSC1/TSC2 with a Defendant-favorable outcome) and drawing an analogy to the input current case.
            \item \textbf{Plaintiff's Rebuttal:} Address and distinguish the counterexample, reinforcing the Plaintiff's original argument.
        \end{enumerate}
        \item Use Provided Factors: Base your arguments on the factors outlined, ensuring logical consistency.
    \end{enumerate}

    \textbf{Example Input and Output}

    \textbf{Example Current Case}
    \begin{itemize}
        \item F1 Disclosure-in-negotiations (D)
        \item F4 Agreed-not-to-disclose (P)
        \item F6 Security-measures (P)
        \item F10 Secrets-disclosed-outsiders (D)
        \item F12 Outsider-disclosures-restricted (P)
        \item F14 Restricted-materials-used (P)
        \item F21 Knew-info-confidential (P)
    \end{itemize}

    \textbf{Example TSC1}
    outcome Plaintiff
    \begin{itemize}
        \item F4 Agreed-not-to-disclose (P)
        \item F6 Security-measures (P)
        \item F7 Brought-tools (P)
        \item F8 Competitive-advantage (P)
        \item F18 Identical-products (P)
    \end{itemize}

    \textbf{Example TSC2}
    outcome Defendant
    \begin{itemize}
        \item F3 Employee-sole-developer (D)
        \item F4 Agreed-not-to-disclose (P)
        \item F5 Agreement-not-specific (D)
        \item F6 Security-measures (P)
        \item F21 Knew-info-confidential (P)
    \end{itemize}

    \textbf{Example Output (json format)}

    \begin{description}
        \item[Plaintiff's Argument:] Factors F4 Agreed-not-to-disclose (P) and F6 Security-measures (P) were present in both the current case and TSC1, where the court found in favor of the Plaintiff. In Addition, Factors F12 Outsider-disclosures-restricted (P), F14 Restricted-materials-used (P), F21 Knew-info-confidential (P) are present in the current case and favor the Plaintiff.
        
        \item[Defendant's Counterargument:] TSC1, cited by the plaintiff is distinguishable because factors F7 Brought-tools (P), F8 Competitive-advantage (P), and F18 Identical-products (P) were also present, but are not present in the current case. In addition, F1 Disclosure-in-negotiations (D) and F10 Secrets-disclosed-outsiders (D) are pro-defendant strengths present in the current case but not in TSC1. TSC2 is a counterexample to TSC1. In TSC2, F4 Agreed-not-to-disclose (P), F6 Security-measures (P), and F21 Knew-info-confidential (P) were present in both the current case and TSC2 and the court found in favor of the Defendant.
        
        \item[Plaintiff's Rebuttal:] TSC2, cited by the Defendant is distinguishable. In TSC2, the additional factors F5 Agreement-not-specific (D) and F3 Employee-sole-developer (D) were present and are not present in the current case. Also, F12 Outsider-disclosures-restricted (P) and F14 Restricted-materials-used (P) are present in the current case but not in TSC2.
    \end{description}

    \textbf{Current Case, TSC1, and TSC2} ...\\

\end{custombox}

\section{Prompt Structure for Factor Extraction}
\label{app:factor_extraction_prompt}
The following shows the structure of the prompt provided to the LLM (\texttt{GPT-4.1}) for the factor extraction task.

\begin{custombox}{Example Prompt}
    \label{prompt:factor_extraction_appendix} 
    \textbf{TASK}\\
    You are tasked with extracting factors from the 3-ply argument.

    \textbf{Example Input (json format)}

    \begin{description}
        \item[Plaintiff's Argument:] Factors F4 Agreed-not-to-disclose (P) and F6 Security-measures (P) were present in both the current case and TSC1, where the court found in favor of the Plaintiff. In Addition, Factors F12 Outsider-disclosures-restricted (P), F14 Restricted-materials-used (P), F21 Knew-info-confidential (P) are present in the current case and favor the Plaintiff.
        
        \item[Defendant's Counterargument:] TSC1, cited by the plaintiff is distinguishable because factors F7 Brought-tools (P), F8 Competitive-advantage (P), and F18 Identical-products (P) were also present, but are not present in the current case. In addition, F1 Disclosure-in-negotiations (D) and F10 Secrets-disclosed-outsiders (D) are pro-defendant strengths present in the current case but not in TSC1. TSC2 is a counterexample to TSC1. In TSC2, F4 Agreed-not-to-disclose (P), F6 Security-measures (P), and F21 Knew-info-confidential (P) were present in both the current case and TSC2 and the court found in favor of the Defendant.
        
        \item[Plaintiff's Rebuttal:] TSC2, cited by the Defendant is distinguishable. In TSC2, the additional factors F5 Agreement-not-specific (D) and F3 Employee-sole-developer (D) were present and are not present in the current case. Also, F12 Outsider-disclosures-restricted (P) and F14 Restricted-materials-used (P) are present in the current case but not in TSC2.
    \end{description}

    \textbf{Example Output (json format)}

    \textbf{Current Case}
    \begin{itemize}
        \item F1 Disclosure-in-negotiations (D)
        \item F4 Agreed-not-to-disclose (P)
        \item F6 Security-measures (P)
        \item F10 Secrets-disclosed-outsiders (D)
        \item F12 Outsider-disclosures-restricted (P)
        \item F14 Restricted-materials-used (P)
        \item F21 Knew-info-confidential (P)
    \end{itemize}

    \textbf{TSC1}
    \begin{itemize}
        \item F4 Agreed-not-to-disclose (P)
        \item F6 Security-measures (P)
        \item F7 Brought-tools (P)
        \item F8 Competitive-advantage (P)
        \item F18 Identical-products (P)
    \end{itemize}

    \textbf{TSC2}
    \begin{itemize}
        \item F3 Employee-sole-developer (D)
        \item F4 Agreed-not-to-disclose (P)
        \item F5 Agreement-not-specific (D)
        \item F6 Security-measures (P)
        \item F21 Knew-info-confidential (P)
    \end{itemize}

    \textbf{3-Ply Argument to be Extracted} ...\\

\end{custombox}


\end{document}